\documentclass[journal]{IEEEtran}

\ifCLASSINFOpdf
\else
  \usepackage[dvips]{graphicx}
  \graphicspath{{../eps/}}
 \fi
\begin{document}

\title{Generative-Adversarial-Networks-based Ghost Recognition}

\author{Yuchen~He,
        Yibing~Chen,
        Sheng~Luo,
        Hui~Chen,
        Jianxing~Li,
        and~Zhuo~Xu% <-this % stops a space
\thanks{Manuscript received Sept 4, 2021. This work was supported in part by the National Natural Science Foundation of China (Grant No. 61901353), in part by the Fundamental Research Funds for the Central Universities (No. xjh012019029), in part by the 111 Project of China (Grant No. B14040).(Corresponding authors: Hui Chen; Zhuo Xu.)}% <-this % stops a space
\thanks{Y. He, Y. Chen, S. Luo, H. Chen, Z. Xu, are with School of Electronic Science and Engineering, Xi'an Jiaotong University, Xi'an 710049, China. (email: yuchenhe@xjtu.edu.cn, chicle.chan@stu.xjtu.edu.cn, lspersist@stu.xjtu.edu.cn, chenhui@xjtu.edu.cn, xuzhuo@xjtu.edu.cn).}
\thanks{J. Li is with School of Information and Communications Engineering, Xi'an Jiaotong University, Xi'an 710049, China. (email: jianxingli.china@xjtu.edu.cn).}}% <-this % stops a space
% <-this % stops a space

% The paper headers
%\markboth{IEEE TRANSACTIONS ON MULTIMEDIA,~Vol.~xxx, No.~xxx, XX~XXXX}%
%{He \MakeLowercase{\textit{et al.}}: Generative-Adversarial-Networks-based Ghost Recognition}

\maketitle

% As a general rule, do not put math, special symbols or citations
% in the abstract or keywords.
\begin{abstract}
Nowadays, target recognition technique plays an important role in many fields.
However, the current target image information based methods suffer from the influence of image quality and the time cost of image reconstruction.
In this paper, we propose a novel imaging-free target recognition method combining ghost imaging (GI) and generative adversarial networks (GAN).
Based on the mechanism of GI, a set of random speckles sequence is employed to illuminate target, and a bucket detector without resolution is utilized to receive echo signal.
The bucket signal sequence formed after continuous detections is constructed into a bucket signal array, which is regarded as the sample of GAN.
Then, conditional GAN is used to map bucket signal array and target category.
In practical application, the speckles sequence in training step is employed to illuminate target, and the bucket signal array is input GAN for recognition.
The proposed method can improve the problems caused by conventional recognition methods that based on target image information, and provide a certain turbulence-free ability.
Extensive experiments show that the proposed method achieves promising performance.
\end{abstract}

% Note that keywords are not normally used for peerreview papers.

\IEEEpeerreviewmaketitle

\section{Introduction}
\IEEEPARstart{G}{host} imaging (GI) has been regarded as a new imaging method since it was proposed~\cite{pittman1995optical, abouraddy2001role, bennink2002two, gatti2004ghost, bennink2004quantum, valencia2005two, ferri2005high, basano2006experiment, scarcelli2006can}.
GI utilizes a series of random speckle patterns to illuminate target, and a bucket detector without resolution to collect echo signal containing target information.
The correlation between bucket and patterns can obtain target image.
Due to its lens-less imaging capability, turbulence-free imaging and high detection sensitivity, GI has received lots of attentions in recent years and made some results~\cite{shapiro2008computational, katz2009compressive, meyers2011turbulence, pelliccia2016experimental, he2018ghost}.

Moreover, GI has been extended to microwave band~\cite{li2014radar, zhu2015radar, cheng2017radar, he2018resolution, zhu2018mixed, zhu2020resolution}.

However, the poor trade-off of image quality and imaging time limits the application of GI.
Nowadays, target recognition technique is an important approach, whether for economy or military~\cite{xiao1994radar, der1997probe, ren2006automatic, gronwall2006ground, chen2008spatiotemporal, iftekharuddin2011transformation, guo2013an, garzon-guerrero2013classification, li2016multiple, yang2016moving, karjalainen2019training, bai2019robust, nasrabadi2019deeptarget, jiang2020using, tao2020convolutional, yang2020d2n4, mao2021target, guo2021robust, ngo2021self}.
Conventional recognition technique is mostly based on target image information.
The quality of target image determines the recognition accuracy and time, which restricted by the means of acquisition and processing target information.
Target image information is obtained mainly by image acquisition (direct video or photo) or reconstruction.
Video or photo quality will be seriously degraded in some special environments, and the image reconstruction consumes extra time.
Recently, generative adversarial networks (GAN) was proposed by Goodfield in 2014 as a new framework, which includes generative model and discriminative model~\cite{goodfellow2014generative}.
Generative model is used to acquire the distribution of the input data and generate new data.
Discriminative model is used to discriminate whether the data is real data or generated by the generator.
GAN is one of the methods with high attention in unsupervised learning field.
Compared with other traditional deep learning methods, GAN is a flexible design framework.
When the probability density can not be calculated, the traditional model which depends on the natural interpretation of data can not be learned and applied. However, GAN introduces the training mechanism of internal confrontation, it can approach some objective functions which are not easy to calculate.
Simultaneously, GAN provides a powerful algorithm framework for unsupervised learning.

In this paper, a novel target recognition method combining GI and GAN is proposed, which can recognize multiple categories of targets by bucket signal array, and regardless of state and attitude of targets.
Based on the mechanism and architecture of GI, a set of random speckles sequence are utilized to illuminate target.
In training step, the bucket signals formed by multiple illuminations are constructed into a bucket signal array and used as training set.
Different targets mean different bucket signal arrays, so we input different bucket signal arrays generated by different targets into network for training.
Noteworthy, different sets of speckles sequence represent different bucket signal arrays.
Therefore, in practical application, the same set of speckles sequence in training step is employed to illuminate target.
Then, potential targets will be recognized quickly.
Handwritten target recognition experiments of numbers and letters are carried out to demonstrate the proposed.
Further validation experiments such as target recognition under different attitudes or turbulence-free conditions are demonstrated.
Finally, a physical experiment at distance of 20m is carried out on different targets.
Extensive experiments above show that the proposed method achieves promising performance.

The main contributions of this work can be summarized as follows:

1) To improve the bottleneck problem brought by the traditional target recognition technique that based on target image information, we propose a novel imaging-free target recognition method combining GI and GAN.
On the premise of fixed random speckles sequence, through training based on GAN, the mapping relationship between bucket signal array and target is established.

2) The proposed method can recognize multiple categories of targets, and regardless of the state and attitude of targets.
Extensive experiments show that the proposed method achieves promising performance on different testing data sets.

3) Since the proposed method is based on GI mechanism and architecture, it can provide certain turbulence-free ability.
The method also can achieve accurate recognition of target in the presence of disturbance.

The rest of this paper is organized as follows.
Section \uppercase\expandafter{\romannumeral2} provides a brief survey of related work.
Section \uppercase\expandafter{\romannumeral3} presents a comprehensive introduction to the Generative-Adversarial-Networks-based ghost recognition is provided.
In Section \uppercase\expandafter{\romannumeral4}, the proposed method is compared and verified by extensive experiments.
The paper is concluded in Section \uppercase\expandafter{\romannumeral5}.

\section{Related Work}
\subsection{Ghost Imaging}
GI was first experimentally demonstrated by Pittman and Shih with entangled source~\cite{pittman1995optical}.
In subsequent studies, researchers found that various of sources can achieve GI.
In 2008, computational ghost imaging (CGI) was proposed~\cite{shapiro2008computational}, which provide a practical scheme for GI.
CGI employs calculation instead of measurement to obtain reference signal.
The proposed method is based on the scheme of CGI.
Similarly, CGI utilizes a series of random speckles to illuminate target, and the restoration process of the target image can be expressed as
\begin{equation}
\centering
{g^{\left( 2 \right)}} = \frac{{\left\langle {{I_{bucket}} \cdot I\left( {x,y} \right)} \right\rangle }}{{\left\langle {{I_{bucket}}} \right\rangle \left\langle {I\left( {x,y} \right)} \right\rangle }}
\end{equation}
where ${g^{\left( 2 \right)}}$ denotes the second-order correlation function, ${\left\langle  \cdot  \right\rangle }$ is the ensemble average, ${{I_{bucket}}}$ is the bucket signal and ${I\left( {x,y} \right)}$ is the reference signal.
The above method and their derivations are widely used in GI.
Because the lens and detector with spatial resolution are omitted, GI possesses many advantages and attracts researchers' attention.
However, GI still faces the challenge of low detection signal-to-noise ratio (SNR) and image peak-signal-to-noise ratio (PSNR).
GI can not simultaneously meet the requirements of high quality image result and short imaging time at present.
Consequently, the application scenarios of GI are limited.
Recently, artificial intelligence (AI) methods are increasingly introduced into GI.
GAN, which has attracted much attention in the field of deep learning in recent years, is also considered to develop the application scenarios of GI.

\subsection{Conditional Generative Adversarial Networks}
The basic idea of GAN can be expressed as:
The generator network G generates the forged image according to the random noise, and the discriminator network D judges whether the image is true or false.
The goal of discriminator D is to judge the real image as true as possible and the forged image of generator G as false, so as to improve the accuracy of judging the true and false.
The goal of generator G is to make the forged image look true as much as possible and reduce the accuracy of discriminator D.
When discriminator D cannot judge whether the image is a real image or generated by generator G, the model training is completed.
The objective function of GAN can be expressed as:
\begin{equation}
\centering
\begin{array}{l}
\mathop {\min }\limits_G \mathop {\max }\limits_D V\left( {D,G} \right) = \\
{E_{x \sim {P_{data}}\left( x \right)}}\left[ {\log D\left( x \right)} \right] + {E_{z \sim {P_z}\left( x \right)}}\left[ {\log \left( {1 - D\left( {G\left( z \right)} \right)} \right)} \right]
\end{array}
\end{equation}
where ${x}$ represents the real image, and ${z}$ represents the input of the generator G.
The right side of the above formula is divided into two terms.
The first term indicates the probability of discriminator D judging the real image as true, the closer to 1, the better.
In the second term, $G\left( z \right)$ represents the image generated by generator G, the task of generator G is to make the image generated close to the real image, the better the performance of G is, the greater the $D\left( {G\left( z \right)} \right)$.
The second term represents the probability of discriminator D judging whether the image generated by generator G is a real image, the better the performance of D, the smaller the $D\left( {G\left( z \right)} \right)$, and there is a process of game.

\begin{figure}[h]
\centering
\includegraphics[width = 8.5 cm]{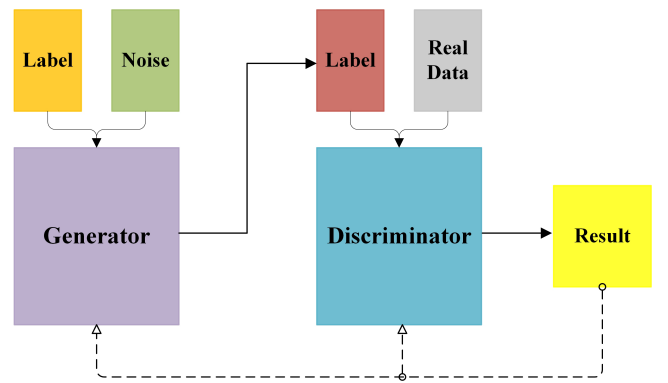}
\caption{Schematic diagram of CGAN.}
\label{CGAN}
\end{figure}

The most important feature of GAN is game competition, so that it does not need to model in advance, and only needs to sample the distribution to approximate the real data.
GAN does not need to rely on a priori assumption like the traditional complex generation model, that is, assuming that the data obey a certain distribution, and then using maximum likelihood to estimate the data distribution.
Although GAN is effective, it also has some disadvantages, it is too free to control the output of the generator.
To solve this problem, Montreal proposes generative adversarial networks with constraints, named conditional generative adversarial networks (CGAN)~\cite{mirza2014conditional}.
In CGAN, conditional variable is added to generator and discriminator to guide the data generation process. CGAN has been proved to be effective and widely used.
Fig.~\ref{CGAN} shows the schematic diagram of CGAN.

The training steps of CGAN can be summarized as follows:

1) Sample real data, obtain the corresponding label y and send it to discriminator, and update the parameters according to the output results.

2) Generate random noise, enter generator with the label y in step (1), and generator generates data.

3) Sent the data generate in step (2) and label y to discriminator.

4) Generator adjusts the parameters according to the output of discriminator.

5) Repeat until generator and discriminator reach Nash equilibrium.

In this paper, CGAN is employed to match bucket signal array and target category.

\section{The Proposed Target Recognition Method}
\subsection{Principle}
\begin{figure*}[!t]
\centering
\includegraphics[width = 18 cm]{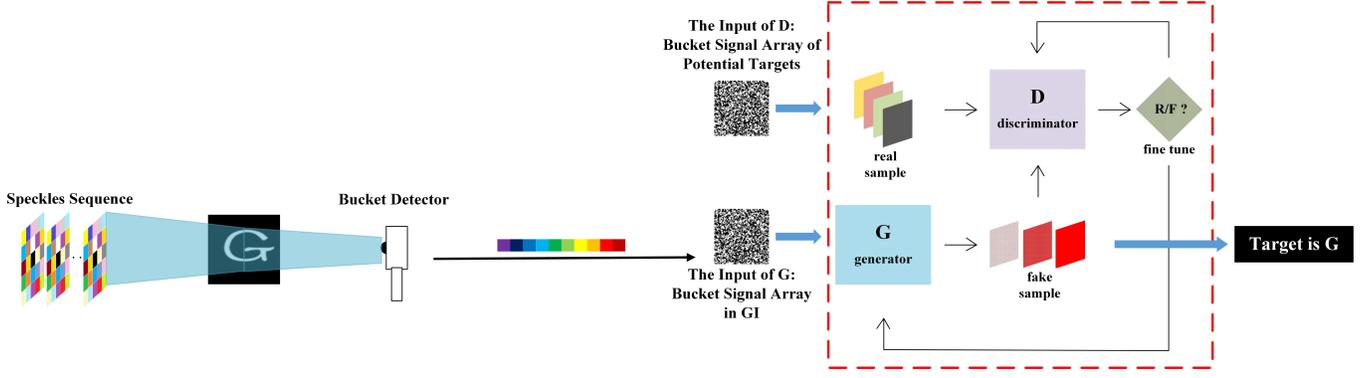}
\caption{Schematic diagram of the proposed method.}
\label{architecture}
\end{figure*}

Combining CGI and CGAN, we propose a imaging-free target recognition technique based on bucket signal array.
Fig.~\ref{architecture} shows the architecture of the proposed method.

As shown in Fig.~\ref{architecture}, a series of different random speckles are utilized to illuminate target respectively, and the echo signal is continuously received by a bucket detector without spatial resolution.
Therefore, what the bucket detector receives is a specific value.
After multiple samplings, we transform this set of signals into a two-dimensional array.
This two-dimensional random array is regarded as one sample of CGAN.
Meanwhile, each two-dimensional bucket signal array represents one label of target.
When changing target, if we still utilize the same set of random speckles to illuminate, different bucket signal array will be generated.
Obviously, different bucket signal arrays represent different target labels.
The bucket signal arrays of different targets are regarded as real samples and sent to discriminator.
Significantly, for same target, one set of random speckles sequence corresponds to one bucket signal array.
Therefore, if the random speckles sequence in training stage is utilized to illuminate target in application, we can recognize the target according to the bucket signal array.

The objective function consists of two parts, log likelihood of real bucket signal array samples ${L_s}$ and log likelihood of real bucket signal class labels ${L_c}$.
%They can be expressed as
%\begin{widetext}
%\begin{equation}
%\centering
%\begin{array}{l}
%{L_s} = E\left[ {\log P\left( {S = real{\rm{|}}{{\rm{X}}_{real}}} \right)} \right] + E\left[ {\log P\left( {S = fake{\rm{|}}{{\rm{X}}_{fake}}} \right)} \right]\\
%{L_c} = E\left[ {\log P\left( {C = c{\rm{|}}{{\rm{X}}_{real}}} \right)} \right] + E\left[ {\log P\left( {C = c{\rm{|}}{{\rm{X}}_{fake}}} \right)} \right]
%\end{array}
%\end{equation}
%\end{widetext}
The goal of discriminator is to find a suitable value to maximize ${L_s} + {L_c}$, and the goal of generator is to find a suitable value to maximize ${L_s} - {L_c}$.
When discriminator can not distinguish whether the input data is real or not, the network reaches equilibrium, and the probability distribution of output is 0.5.
By introducing the bucket signal array class label as a condition variable, the uncontrollable shortcoming of the traditional GAN training process is improved.
Finally, the label control generator of the bucket signal array can generate the corresponding category of bucket signal data.
Simultaneously, the discriminator also learns the data feature distribution of real samples in the process of confrontation to judge whether the input is real or generated.

\subsection{Network Architecture}
\begin{figure}[!t]
\centering
\includegraphics[width = 8.5 cm]{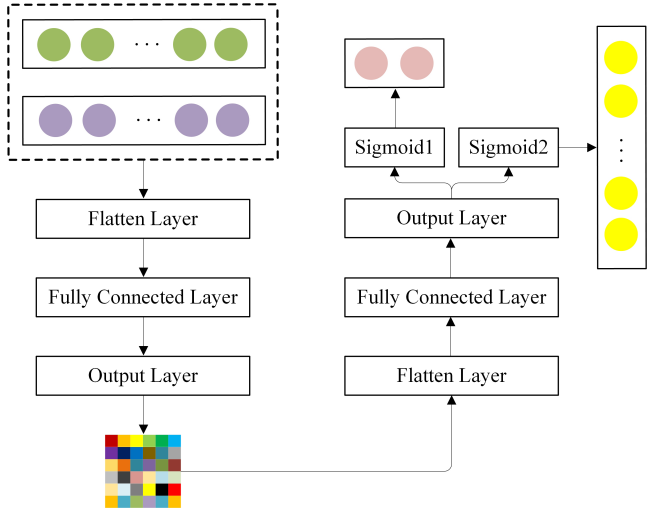}
\caption{Network architecture of the proposed method.}
\label{network}
\end{figure}

In this paper, the construction of CGAN is based on Multilayer Perceptron (MLP), which is a deep artificial neural network.
A MLP neural network generally contains multiple perceptrons, including an input layer to receive input data and an output layer to classify or predict input data. There are usually many hidden layers between the input layer and the output layer.
The calculation process of MLP is completed in the hidden layer.
The schematic diagram of the network architecture of the proposed method is shown in Fig.~\ref{network}, where b is the size of batch and n is the number of bucket signal labels.

The generator network includes seven layers (one input layer, one flatten layer, four fully connected layers, and one output layer), which used to generate images of the same type as the target.
The number of nodes in fully connected layer third to sixth is 256, 512 and 1024 respectively.
The output layer reconstructs the output of the previous layer to generate image, which type is same as the input.
Similarly, the discriminator is based on MLP, the network also includes seven layers (one input layer, one flatten layer, three fully connected layers and two output layers), which used to identify the category of the input image.
In the flatten layer, the input image is tiled into 784 nodes.
The third to fifth layer is fully connected layer, and the number of output nodes is 512.
The sixth and seventh layer are the output layer of the discriminator.
The sixth layer uses sigmoid function to normalize the output of the upper layer, and the output of sixth layer is the authenticity of the training sample image.
In the seventh layer, the output of the upper layer is normalized by softmax function, and the output of seventh layer is the category of the training sample image.
The three fully connected layers in discriminator are normalized, and all the excitation functions use leakyReLU function.
The output of fully connected layers is the probability of target categories corresponding to the input samples and the authenticity of the training samples.
The cross entropy between the output of the training samples and the label is taken as the training loss function.
The output of the training samples is compared with the label, and the optimizer Adam is used to control the change of the learning rate. Finally, the operation to reduce the loss is defined for the confrontation training.

The steps can be summarized as:

1) Prepare training and testing sets. According to CGI mechanism, using a set of random speckles sequence to illuminate potential targets several times, and the bucket signal arrays collected as training set.

2) Build network. Based on tensorflow-gpu 1.13, keras version 2.1.5, using pycharm, three models are built: generator model, discriminator model and confrontation training model.

3) Train network. In the loop, the images in training set and their corresponding categories are input, and the generator and discriminator model are trained simultaneously.

4) Test network. The same speckles sequence as in training is used to illuminate, and the bucket signal array is sent to the network to recognize the target.

\section{Experimental Results}
\subsection{Training Settings}
\begin{figure}[h]
\centering
\includegraphics[width = 8.5 cm]{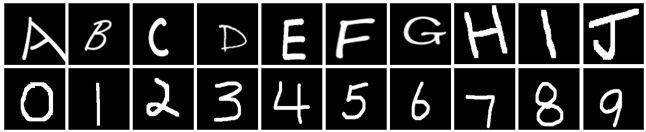}
\caption{Handwritten targets in our experiment.}
\label{target}
\end{figure}

Under the architecture of CGI, the training data comes from bucket signal after multiple samplings of target based on a set of fixed random speckles sequence.
The size of speckle and target is both 28*28.
Each pixel of the target is sampled once, and 784 bucket signal values are formed.
We construct the bucket signal sequence into a bucket signal array as an input of CGAN.
Consequently, the size of bucket signal array is 28*28.
In the following description, each input of CGAN is formed after a complete round of sampling, that is, 784.
The targets in our experiment are handwritten letters and numbers.
Letter targets include 10 categories of "A,B,C,D,E,F,G,H,I,J", and number targets include 10 categories of "0,1,2,3,4,5,6,7,8,9".
We trained four networks using 5000, 10000, 20000 and 60000 samples, and each category of target contains 500, 1000, 2000 and 6000 in four networks, respectively.
Meanwhile, 500, 1000, 2000 and 5000 epochs are performed in each network. Fig.~\ref{target} shows the handwritten targets in our experiment.

\subsection{Results on Letters and Numbers Targets}
\begin{figure}[h]
\centering
\includegraphics[width = 8.5 cm]{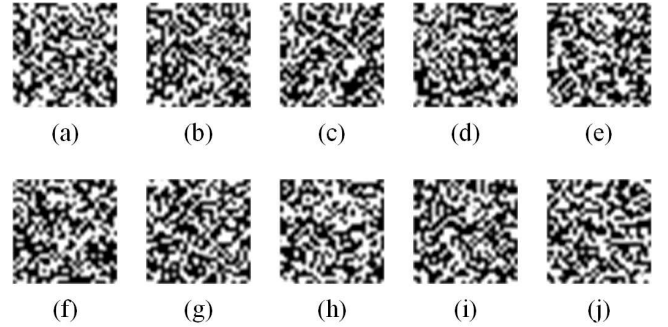}
\caption{Bucket signal arrays of different kinds of letter targets. (a) A. (b) B. (c) C. (d) D. (e) E. (f) F. (g) G. (h) H. (i) I. (j) J.}
\label{bucket-let}
\end{figure}

Firstly, we test the proposed method on the letter handwriting targets set.
No matter which method of target recognition, it is based on the different characteristics of the target to distinguish and complete the recognition, the proposed method is no exception.
We convert the received bucket signals sequence into array, which is used as training and testing samples.
Fig.~\ref{bucket-let} shows the bucket signal arrays for 10 categories of targets in letter experiment.

These 10 categories in Fig.~\ref{bucket-let} respectively represent the bucket signal array formed by 10 letter targets after 784 times of illumination.
Obviously, different categories of targets  have different forms of bucket signal arrays.
Then, we compare the correlation between each two bucket signal arrays for letter targets for quantitative.
\begin{table}[!t]
\caption{Correlation Coefficient Between Bucket Signal Arrays For Letter Targets.}
\label{Correlation-letter}
\centering
\begin{tabular}{|c|c|c|c|c|}
\hline
B0-B1  & B0-B2  & B0-B3  & B0-B4  & B0-B5  \\ \hline
0.1153 & 0.2008 & 0.3155 & 0.1529 & 0.2754 \\ \hline
B0-B6  & B0-B7  & B0-B8  & B0-B9  & B1-B2  \\ \hline
0.2101 & 0.1779 & 0.4078 & 0.3195 & 0.2003 \\ \hline
B1-B3  & B1-B4  & B1-B5  & B1-B6  & B1-B7  \\ \hline
0.2037 & 0.0595 & 0.1702 & 0.2906 & 0.1057 \\ \hline
B1-B8  & B1-B9  & B2-B3  & B2-B4  & B2-B5  \\ \hline
0.1787 & 0.2167 & 0.2406 & 0.1896 & 0.1364 \\ \hline
B2-B6  & B2-B7  & B2-B8  & B2-B9  & B3-B4  \\ \hline
0.2663 & 0.0676 & 0.2801 & 0.2306 & 0.1198 \\ \hline
B3-B5  & B3-B6  & B3-B7  & B3-B8  & B3-B9  \\ \hline
0.3928 & 0.388  & 0.226  & 0.3525 & 0.1786 \\ \hline
B4-B5  & B4-B6  & B4-B7  & B4-B8  & B4-B9  \\ \hline
0.1831 & 0.1819 & 0.2436 & 0.2277 & 0.1579 \\ \hline
B5-B6  & B5-B7  & B5-B8  & B5-B9  & B6-B7  \\ \hline
0.2998 & 0.1768 & 0.3192 & 0.2694 & 0.2013 \\ \hline
B6-B8  & B6-B9  & B7-B8  & B7-B9  & B8-B9  \\ \hline
0.366  & 0.2358 & 0.1728 & 0.3734 & 0.2238 \\ \hline
\end{tabular}
\end{table}

Table~\ref{Correlation-letter} shows the correlation coefficient between bucket signal arrays of different target.
We can conclude that the correlation between these bucket signal arrays is low, and we can train the network according to these differences.
Then, we randomly select 10 samples from test set (containing 100 samples that not in the training set) for testing.
The recognition results are shown in Table~\ref{Results-Letter}.

\begin{table*}[!t]
\caption{Recognition Results For Letter Targets.}
\label{Results-Letter}
\centering
\begin{tabular}{|c|c|c|c|c|c|c|c|c|c|c|c|c|c|c|c|c|}
\hline
Samples   & \multicolumn{4}{c|}{5000}              & \multicolumn{4}{c|}{10000}             & \multicolumn{4}{c|}{20000}             & \multicolumn{4}{c|}{60000}             \\ \hline
Epoch     & 500   & 1000  & 2000  & 5000           & 500   & 1000  & 2000  & 5000           & 500   & 1000  & 2000  & 5000           & 500   & 1000  & 2000  & 5000           \\ \hline
Target\_A & 90\%  & 100\% & 100\% & \textbf{100\%} & 100\% & 100\% & 100\% & \textbf{100\%} & 100\% & 70\%  & 90\%  & \textbf{100\%} & 100\% & 100\% & 80\%  & \textbf{100\%} \\ \hline
Target\_B & 80\%  & 100\% & 100\% & \textbf{100\%} & 100\% & 100\% & 100\% & \textbf{100\%} & 80\%  & 100\% & 100\% & \textbf{100\%} & 90\%  & 100\% & 100\% & \textbf{100\%} \\ \hline
Target\_C & 100\% & 100\% & 100\% & \textbf{80\%}  & 70\%  & 90\%  & 100\% & \textbf{100\%} & 100\% & 90\%  & 90\%  & \textbf{100\%} & 80\%  & 100\% & 100\% & \textbf{100\%} \\ \hline
Target\_D & 100\% & 100\% & 90\%  & \textbf{100\%} & 100\% & 100\% & 100\% & \textbf{100\%} & 90\%  & 100\% & 100\% & \textbf{100\%} & 100\% & 100\% & 90\%  & \textbf{100\%} \\ \hline
Target\_E & 100\% & 100\% & 100\% & \textbf{100\%} & 90\%  & 100\% & 100\% & \textbf{90\%}  & 100\% & 90\%  & 100\% & \textbf{100\%} & 100\% & 100\% & 80\%  & \textbf{100\%} \\ \hline
Target\_F & 100\% & 100\% & 100\% & \textbf{80\%}  & 100\% & 100\% & 100\% & \textbf{100\%} & 80\%  & 100\% & 100\% & \textbf{100\%} & 90\%  & 100\% & 100\% & \textbf{100\%} \\ \hline
Target\_G & 100\% & 100\% & 90\%  & \textbf{100\%} & 100\% & 100\% & 100\% & \textbf{100\%} & 80\%  & 100\% & 90\%  & \textbf{100\%} & 100\% & 90\%  & 100\% & \textbf{100\%} \\ \hline
Target\_H & 100\% & 90\%  & 100\% & \textbf{100\%} & 100\% & 100\% & 100\% & \textbf{100\%} & 80\%  & 90\%  & 100\% & \textbf{100\%} & 70\%  & 100\% & 100\% & \textbf{100\%} \\ \hline
Target\_I & 100\% & 80\%  & 100\% & \textbf{100\%} & 100\% & 100\% & 100\% & \textbf{100\%} & 90\%  & 100\% & 100\% & \textbf{100\%} & 100\% & 100\% & 100\% & \textbf{100\%} \\ \hline
Target\_J & 90\%  & 90\%  & 100\% & \textbf{100\%} & 100\% & 100\% & 100\% & \textbf{100\%} & 100\% & 100\% & 100\% & \textbf{100\%} & 100\% & 80\%  & 100\% & \textbf{100\%} \\ \hline
\end{tabular}
\end{table*}

\begin{figure}[h]
\centering
\includegraphics[width = 8.5 cm]{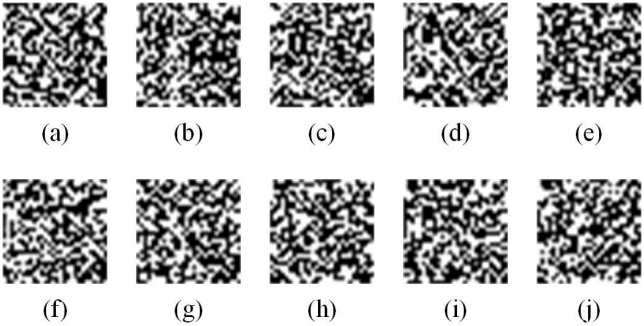}
\caption{Bucket signal arrays of different kinds of number targets. (a) 0. (b) 1. (c) 2. (d) 3. (e) 4. (f) 5. (g) 6. (h) 7. (i) 8. (j) 9.}
\label{bucket-num}
\end{figure}

Table~\ref{Results-Letter} shows that with the increase of the number of samples, the recognition accuracy eventually reaches an ideal level.
Similarly with letter targets, Fig.~\ref{bucket-num} shows the bucket signal arrays for 10 categories of targets in number experiment.

As can be seen from Fig.~\ref{bucket-num}, the arrays formed by the bucket signals sequence of different targets are different.
Based on this, we test the proposed method on four networks with different sizes and epochs, and Table~\ref{Results-number} shows the results.

\begin{table*}[!t]
\caption{Recognition Results For For Number Targets.}
\label{Results-number}
\centering
\begin{tabular}{|c|c|c|c|c|c|c|c|c|c|c|c|c|c|c|c|c|}
\hline
Samples   & \multicolumn{4}{c|}{5000}              & \multicolumn{4}{c|}{10000}             & \multicolumn{4}{c|}{20000}             & \multicolumn{4}{c|}{60000}             \\ \hline
Epoch     & 500   & 1000  & 2000  & 5000           & 500   & 1000  & 2000  & 5000           & 500   & 1000  & 2000  & 5000           & 500   & 1000  & 2000  & 5000           \\ \hline
Target\_0 & 90\%  & 100\% & 100\% & \textbf{100\%} & 60\%  & 100\% & 100\% & \textbf{100\%} & 100\% & 70\%  & 100\% & \textbf{100\%} & 100\% & 100\% & 100\% & \textbf{100\%} \\ \hline
Target\_1 & 100\% & 70\%  & 100\% & \textbf{100\%} & 100\% & 80\%  & 100\% & \textbf{100\%} & 100\% & 100\% & 70\%  & \textbf{100\%} & 90\%  & 100\% & 100\% & \textbf{90\%}  \\ \hline
Target\_2 & 90\%  & 100\% & 100\% & \textbf{100\%} & 100\% & 100\% & 100\% & \textbf{90\%}  & 100\% & 100\% & 100\% & \textbf{100\%} & 70\%  & 100\% & 100\% & \textbf{100\%} \\ \hline
Target\_3 & 100\% & 80\%  & 100\% & \textbf{100\%} & 100\% & 100\% & 80\%  & \textbf{100\%} & 100\% & 100\% & 100\% & \textbf{100\%} & 100\% & 90\%  & 100\% & \textbf{100\%} \\ \hline
Target\_4 & 100\% & 100\% & 100\% & \textbf{100\%} & 100\% & 100\% & 90\%  & \textbf{90\%}  & 100\% & 80\%  & 100\% & \textbf{100\%} & 60\%  & 100\% & 80\%  & \textbf{100\%} \\ \hline
Target\_5 & 100\% & 100\% & 100\% & \textbf{100\%} & 100\% & 100\% & 100\% & \textbf{100\%} & 100\% & 100\% & 100\% & \textbf{100\%} & 100\% & 100\% & 100\% & \textbf{100\%} \\ \hline
Target\_6 & 100\% & 90\%  & 100\% & \textbf{100\%} & 100\% & 100\% & 100\% & \textbf{100\%} & 100\% & 70\%  & 100\% & \textbf{100\%} & 100\% & 100\% & 100\% & \textbf{100\%} \\ \hline
Target\_7 & 90\%  & 100\% & 80\%  & \textbf{90\%}  & 90\%  & 100\% & 100\% & \textbf{100\%} & 90\%  & 80\%  & 100\% & \textbf{100\%} & 100\% & 70\%  & 80\%  & \textbf{100\%} \\ \hline
Target\_8 & 100\% & 100\% & 100\% & \textbf{100\%} & 100\% & 100\% & 100\% & \textbf{100\%} & 100\% & 100\% & 100\% & \textbf{100\%} & 100\% & 100\% & 100\% & \textbf{100\%} \\ \hline
Target\_9 & 90\%  & 100\% & 100\% & \textbf{100\%} & 100\% & 100\% & 100\% & \textbf{100\%} & 100\% & 90\%  & 100\% & \textbf{100\%} & 100\% & 100\% & 90\%  & \textbf{100\%} \\ \hline
\end{tabular}
\end{table*}

\subsection{Same Random Speckles Sequence}
\begin{figure}[h]
\centering
\includegraphics[width = 8.5 cm]{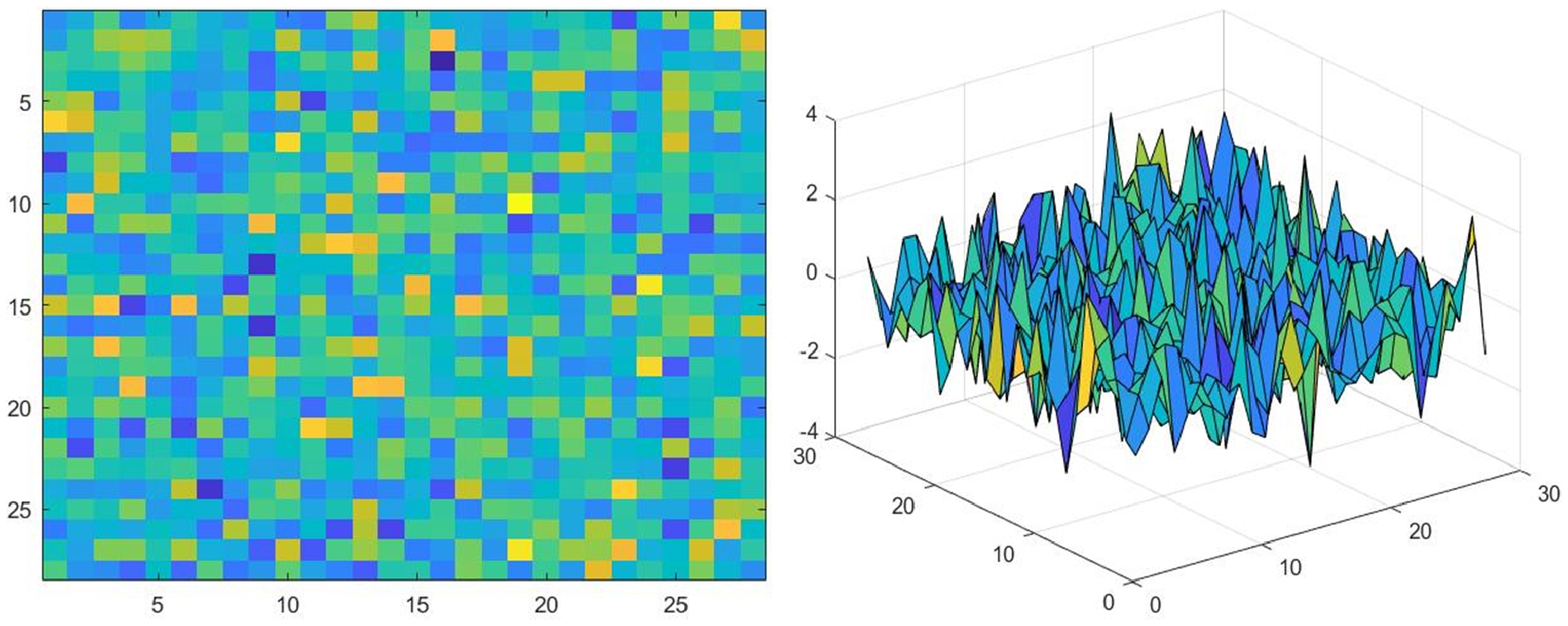}
\caption{Speckle.}
\label{speckles}
\end{figure}

In the experiments above, two different speckle sequences are used for training and testing on letter and number targets.
In this section, we try to train letters and numbers targets with one set of random speckle sequence, and so do the test.
Similarly, the size of one speckle is 28*28, and these 784 random speckles are different and spatial independence.
We randomly select one random speckle and show in Fig.~\ref{speckles}.

Then, the bucket signal array formed by 784 illuminations is taken as one sample.
We use 5000 and 10000 samples to train two networks, each with 2000 and 5000 epochs, respectively.
Table~\ref{Letter and Number} lists the testing results.

\begin{table*}[!t]
\caption{Testing Results on Letter and Number Targets with Same Speckles Sequence.}
\label{Letter and Number}
\centering
\begin{tabular}{|c|c|c|c|c|c|c|c|c|c|}
\hline
Samples   & \multicolumn{2}{c|}{5000} & \multicolumn{2}{c|}{10000} & Samples   & \multicolumn{2}{c|}{5000} & \multicolumn{2}{c|}{10000} \\ \hline
Epoch     & 2000        & 5000        & 2000         & 5000        & Epoch     & 2000        & 5000        & 2000         & 5000        \\ \hline
Target\_A & 100\%       & 100\%       & 100\%        & 100\%       & Target\_0 & 90\%        & 90\%        & 100\%        & 100\%       \\ \hline
Target\_B & 100\%       & 100\%       & 80\%         & 100\%       & Target\_1 & 100\%       & 50\%        & 100\%        & 100\%       \\ \hline
Target\_C & 80\%        & 100\%       & 60\%         & 100\%       & Target\_2 & 100\%       & 100\%       & 100\%        & 100\%       \\ \hline
Target\_D & 100\%       & 90\%        & 100\%        & 100\%       & Target\_3 & 100\%       & 100\%       & 100\%        & 100\%       \\ \hline
Target\_E & 90\%        & 100\%       & 100\%        & 90\%        & Target\_4 & 100\%       & 100\%       & 100\%        & 100\%       \\ \hline
Target\_F & 100\%       & 80\%        & 100\%        & 100\%       & Target\_5 & 70\%        & 100\%       & 100\%        & 100\%       \\ \hline
Target\_G & 60\%        & 100\%       & 100\%        & 100\%       & Target\_6 & 100\%       & 100\%       & 70\%         & 100\%       \\ \hline
Target\_H & 100\%       & 100\%       & 100\%        & 100\%       & Target\_7 & 100\%       & 60\%        & 100\%        & 90\%        \\ \hline
Target\_I & 100\%       & 100\%       & 100\%        & 100\%       & Target\_8 & 100\%       & 100\%       & 100\%        & 100\%       \\ \hline
Target\_J & 100\%       & 80\%        & 100\%        & 100\%       & Target\_9 & 100\%       & 100\%       & 100\%        & 100\%       \\ \hline
\end{tabular}
\end{table*}

Table~\ref{Letter and Number} shows that the proposed method achieves promising performance.

\subsection{Results on Different Attitudes}
\begin{figure}[h]
\centering
\includegraphics[width = 8.5 cm]{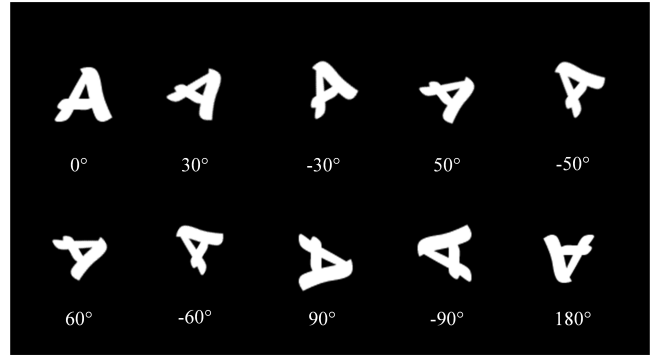}
\caption{Attitudes.}
\label{attitudes}
\end{figure}

In practical applications, the attitude and position of objects are often changing.
Target needs to be successfully recognized in different attitudes.
Therefore, we try this task with the proposed method.
Fig.~\ref{attitudes} shows the different attitudes of letter target A.

We test the recognition performance of 10 different attitudes, including 0$^\circ$, 30$^\circ$, -30$^\circ$, 50$^\circ$, -50$^\circ$, 60$^\circ$, -60$^\circ$, 90$^\circ$, -90$^\circ$, 180$^\circ$.
Table~\ref{Attitudes} shows the results of different attitudes of letter target A.

\begin{table}[h]
\caption{Testing Results on Different Attitudes of Target A.}
\label{Attitudes}
\centering
\begin{tabular}{|c|c|c|c|c|}
\hline
Samples       & \multicolumn{2}{c|}{10000} & \multicolumn{2}{c|}{20000} \\ \hline
Epoch         & 2000         & 5000        & 2000         & 5000        \\ \hline
Attitude\_0   & 100\%        & 90\%        & 100\%        & 100\%       \\ \hline
Attitude\_30  & 100\%        & 100\%       & 100\%        & 100\%       \\ \hline
Attitude\_-30 & 100\%        & 100\%       & 80\%         & 100\%       \\ \hline
Attitude\_50  & 100\%        & 100\%       & 100\%        & 100\%       \\ \hline
Attitude\_-50 & 100\%        & 100\%       & 100\%        & 100\%       \\ \hline
Attitude\_60  & 100\%        & 100\%       & 100\%        & 100\%       \\ \hline
Attitude\_-60 & 100\%        & 100\%       & 100\%        & 100\%       \\ \hline
Attitude\_90  & 100\%        & 100\%       & 100\%        & 100\%       \\ \hline
Attitude\_-90 & 100\%        & 100\%       & 100\%        & 90\%        \\ \hline
Attitude\_180 & 100\%        & 100\%       & 80\%         & 100\%       \\ \hline
\end{tabular}
\end{table}

Table~\ref{Attitudes} shows that regardless the target attitude, the recognition rate of the proposed method is considerable.

\subsection{Turbulence-free Experiment}
Traditional recognition technology based on target image information will seriously affect the accuracy of recognition when there is turbulence disturbance in the scene.
GI is based on correlation detection, so it has certain turbulence-free characteristics.
The imaging process expression of GI can be expressed as:

\begin{equation}
\centering
{B_{bucket}} = {S_{speckle}} \cdot {T_{target}}
\end{equation}
where ${B_{bucket}}$ represents bucket signal, ${S_{speckle}}$ represents speckle sequence and ${T_{target}}$ represents target information.
The speckle sequence is known and the bucket signal is received, so we can obtain the information of target.
When there is turbulence, ${S_{speckle}}$ is polluted, which is different from the projected speckle information.
The imaging process becomes:

\begin{equation}
\centering
B_{bucket}^{'} = S_{speckle}^{'}\left( t \right) \cdot {T_{target}}
\end{equation}
where $S_{speckle}^{'}\left( t \right)$ represents the polluted speckles which change over time, and we can not obtain it accurately.
Consequently, the bucket signal is also polluted and becomes $B_{bucket}^{'}$.
The mapping between ${B_{bucket}}$ and target category is destroyed.
However,  when recognition time is less than the change rate of turbulence, we can think of turbulence as unchangeable.
If the relationship between $B_{bucket}^{'}$ and target category can be mapped through training, we can recognize the target.

The 'XJTU' of handwriting are set as target.
\begin{figure}[h]
\centering
\includegraphics[width = 5 cm]{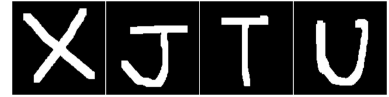}
\caption{'XJTU' target.}
\label{target-xjtu}
\end{figure}

We add a set of fixed random noise (The size is 28*28) to bucket signal array of each category of targets.
The polluted bucket signal arrays are used as training samples.
We employ 4000 and 8000 samples to train two networks, respectively, and each network contains 2000 and 5000 epochs.
Experiment results are shown in Table~\ref{target-xjtu}.
The results show that the proposed method can achieve target recognition in the presence of noise.

\begin{table}[h]
\caption{Testing Results on Turbulence-free of Target XJTU.}
\label{target-xjtu}
\centering
\begin{tabular}{|c|c|c|c|c|}
\hline
Samples   & \multicolumn{2}{c|}{4000} & \multicolumn{2}{c|}{8000} \\ \hline
Epoch     & 2000    & 5000            & 2000    & 5000            \\ \hline
Target\_X & 100\%   & \textbf{100\%}  & 100\%   & \textbf{100\%}  \\ \hline
Target\_J & 100\%   & \textbf{100\%}  & 100\%   & \textbf{100\%}  \\ \hline
Target\_T & 80\%    & \textbf{100\%}  & 100\%   & \textbf{100\%}  \\ \hline
Target\_U & 90\%    & \textbf{100\%}  & 100\%   & \textbf{100\%}  \\ \hline
\end{tabular}
\end{table}

Furthermore, we test the recognition ability of the proposed method under fixed noise level.
Different from random noise, the training samples of fixed noise level are bucket signal arrays without noise.
In testing, we add different fixed levels of noise to the bucket signal array.
We carry out this experiments under 9 different levels of SNR, 14dB, 8dB, 4dB, 2dB, 0dB, -1dB, -3dB, -4dB and -5dB, respectively.
The target is also handwriting of 'XJTU'.
The training set contains 6000 samples and the epoch is 5000.

\begin{table*}[!t]
\caption{Testing Results on Different Levels of SNR.}
\label{target1-xjtu}
\centering
\begin{tabular}{|c|c|c|c|c|c|c|c|c|c|}
\hline
SNR       & 14dB  & 8dB  & 4dB  & 2dB   & 0dB   & -1dB   & -3dB   & -4dB  & -5dB  \\ \hline
Target\_X & 100\% & 100\% & 100\% & 100\% & 100\% & 100\% & 100\% & 90\% & 70\% \\ \hline
Target\_J & 100\% & 100\% & 100\% & 70\%  & 10\%  & 10\%  & 0\%   & 0\%  & 0\%  \\ \hline
Target\_T & 100\% & 90\%  & 80\%  & 60\%  & 60\%  & 30\%  & 10\%  & 10\% & 0\%  \\ \hline
Target\_U & 100\% & 100\% & 100\% & 100\% & 100\% & 100\% & 80\%  & 60\% & 50\% \\ \hline
\end{tabular}
\end{table*}

Table~\ref{target1-xjtu} show that complex fonts, such as 'X' and 'U', have higher recognition accuracy when SNR is low.
Relatively, for simple fonts, such as 'J' and 'T', the accuracy of recognition decreases with the decrease of SNR evidently.

\subsection{Physical Experiment}
\begin{figure}[h]
\centering
\includegraphics[width = 9 cm]{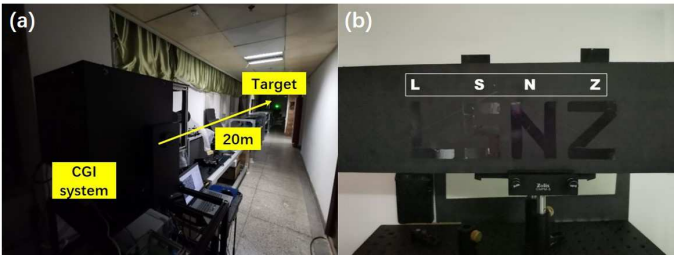}
\caption{Physical experiment. (a) Experimental light path. (b) Target 'LSNZ'.}
\label{physicale}
\end{figure}

Finally, we demonstrated a physical experiment at a distance of 20m.
The physical experiment is based on the CGI architecture, and the targets are reflective letters 'LSNZ'.
Fig.~\ref{physicale} shows the physical experimental light path.

In training, a 10*10 bucket signal array (one sample) is formed through 100 detections for each target.
The training set contains 24000 samples (each target contains 6000 samples), and we trained 1000, 6000 and 8000 epochs.
In testing, the speckles sequence in training step is employed to illuminate targets.
The bucket signal arrays (10*10) are input the trained CGAN for recognition at the same time.
The recognition results are shown in the Table~\ref{physical}.

\begin{table*}[!t]
\caption{Physical experimental results.}
\label{physical}
\centering
\begin{tabular}{|c|c|c|c|}
\hline
           & \multicolumn{3}{c|}{Epoch} \\ \hline
Target    & 1000   & 4000    & 8000    \\ \hline
Target\_L & 60\%   & 70\%    & 100\%   \\ \hline
Target\_S & 90\%   & 100\%   & 100\%   \\ \hline
Target\_N & 50\%   & 80\%    & 100\%   \\ \hline
Target\_Z & 90\%   & 100\%   & 100\%   \\ \hline
\end{tabular}
\end{table*}

Table~\ref{physical} shows that the proposed method can also recognize different targets in physical experiment, and the recognition rate becomes better with the increase of epoch.
It took 0.12s to identify 4 targets, with an average of 0.03s for one target.

\section{Conclusion}
In this paper, we proposed a novel imaging-free target recognition method, which based on GI mechanism and combined with the approach of GAN.
We utilize a set of random speckles sequence to illuminate target, and one bucket detector is employed to received echo signal continuously.
Then, the bucket signal sequence is constructed into a bucket array, which is regarded as one sample of GAN.
After multiple illuminations using this set of speckle sequences, the training sample set is formed.
Meanwhile, we use CGAN method to mapping target category and bucket signal array.
In practical application, the set of speckles sequence in training is still employed to illuminate target, and the bucket signal array is input GAN for recognition. Extensive experiments show that the proposed method achieves promising performance on letters testing data set, numbers testing data set and different attitudes testing data set.
Moreover, the proposed method can provide a certain turbulence-free ability.
Finally, a physical experiment at distance of 20m is carried out to demonstrate the proposed method on different targets.
The proposed method can improve the impact on recognition speed and accuracy, which caused by image acquisition means and quality in the traditional recognition method based on target image information.

\bibliographystyle{IEEEtran}
\bibliography{TMM}
%

%
%% that's all folks
\end{document}